%% file: root.tex
\documentclass[letterpaper, 10pt, conference]{ieeeconf}  

\IEEEoverridecommandlockouts    
\pdfoutput=1

\usepackage{graphics}           
\usepackage{epsfig}             
\usepackage{times}              
\usepackage{amsmath}            
\usepackage{amssymb}            
\usepackage{balance}            
\usepackage{cite}               
\usepackage{graphicx}           
\usepackage{physics}            
\usepackage{xcolor}             
\usepackage{tikz}               

\usepackage{hyperref}           
\usepackage{caption}            
\usepackage{subcaption}         
\usepackage{lipsum}             
\usepackage{booktabs}           
\usepackage{tabularx}           
\usepackage{wrapfig}            
\usepackage{multirow}           

\title{\LARGE \bf
FILIC: Dual-Loop Force-Guided Imitation Learning with Impedance Torque Control for Contact-Rich Manipulation Tasks
}

\author{
Haizhou Ge$^{1*}$, Yufei Jia$^{1*}$, Zheng Li$^{2*}$, Yue Li$^{3*}$, Zhixing Chen$^{1*}$,\\
Lu Shi$^{1\dag}$, Lei Han$^{3\dag}$, Ruqi Huang$^{1\dag}$, Guyue Zhou$^{1\dag}$
\thanks{*Equal contribution; \dag Corresponding Author.}
\thanks{$^{1}$Tsinghua University.
        {\tt\small \{ghz23,jyf23,chenzx24\}@mails.
        tsinghua.edu.cn,
        ruqihuang@sz.tsinghua.edu.cn,
        \{shilu,zhouguyue\}@air.tsinghua.edu.cn}}
\thanks{$^{2}$The Hong Kong University of Science and Technology (Guangzhou).
        {\tt\small zli514@connect.hkust-gz.edu.cn}}
\thanks{$^{3}$DISCOVER Robotics.
        {\tt\small \{lue,leo\}@discover-robotics.com}}
}

\begin{document}
\let\oldtwocolumn\twocolumn
\renewcommand\twocolumn[1][]{%
    \oldtwocolumn[{#1}{
    \vspace{-5pt}
    \begin{center}
        \includegraphics[width=\textwidth]{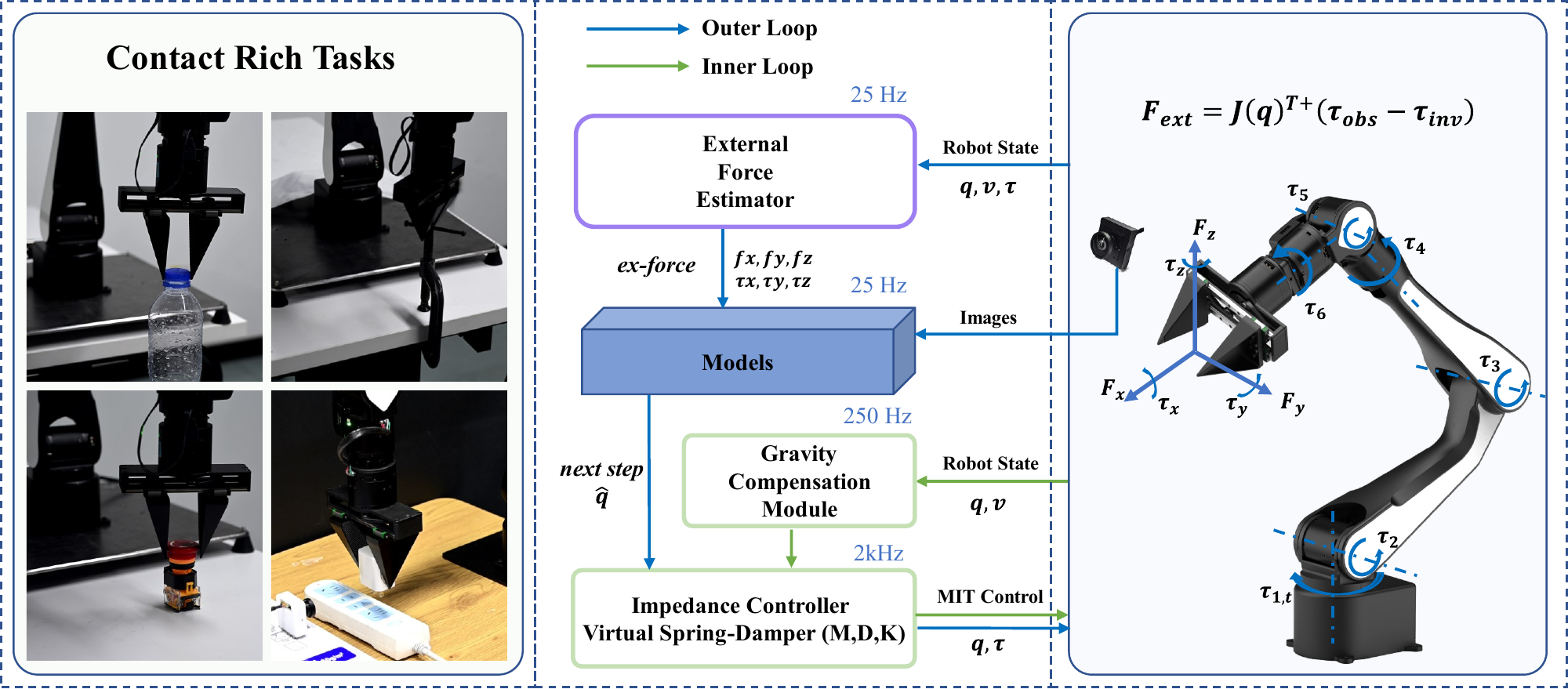}
        \captionof{figure}{\textbf{FILIC}. \textbf{(Left)} Examples of contact-rich manipulation tasks. \textbf{(Middle)} Dual-loop architecture: an outer-loop imitation learning model processes multimodal sensory inputs (estimated external force from an estimator and visual images from cameras) to predict the next action at 25 Hz; a gravity compensation module ensures accurate dynamics estimation at 250 Hz; the inner-loop impedance controller runs at 2 kHz for compliant torque control. \textbf{(Right)} Illustration of external forces estimation in joint space which are computed from joint torques using the robot's Jacobian.}
        \label{fig:teaser}
    \end{center}
    \vspace{0pt}
    }]
}

\maketitle


\input{Sections/0_Abstract}
\input{Sections/1_Introduction}
\input{Sections/2_RelatedWorks}
\input{Sections/3_Preliminaries}

\input{Sections/4_Methodology}
\input{Sections/5_Experiments}
\input{Sections/6_Conclusion}

\input{Sections/7_Bibliography}

\end{document}

%% file: Sections/0_Abstract.tex
\begin{abstract}

Many contact-rich manipulation tasks require precise force regulation. However, most imitation learning (IL) policies remain position-centric and lack explicit force awareness, and adding force/torque sensors to collaborative robot arms is often costly and requires additional hardware design. To overcome these issues, we propose FILIC, a Force-guided Imitation Learning framework with impedance torque control. FILIC integrates a Transformer-based IL policy with an impedance controller in a dual-loop structure, enabling compliant force-informed, force-executed manipulation. For robots without force/torque sensors, we introduce a cost-effective end-effector force estimator using joint torque measurements through analytical Jacobian-based inversion while compensating with model-predicted torques from a digital twin. Experiments show that FILIC significantly outperforms vision-only and joint-torque-based methods, achieving safer, more compliant, and adaptable contact-rich manipulation. The source code is available at \url{https://github.com/OpenGHz/mujoco-wrench-estimator.git}.

\end{abstract}

%% file: Sections/1_Introduction.tex
\section{Introduction} \label{sec 1}

Contact-rich manipulation, requiring precise force control, is fundamental for robots to interact effectively with physical environments. It underpins critical tasks such as insertion~\cite{zhang2025ta}, assembly~\cite{assembly}, and in-hand manipulation~\cite{inhand}, serving as a cornerstone for achieving human-like dexterity in unstructured settings. Imitation learning (IL) has emerged as a promising paradigm in this domain, enabling robots to acquire complex skills from human demonstrations and adapt efficiently to diverse contact scenarios~\cite{imicota}.
Despite recent advancements, significant barriers remain. Conventional IL methods typically output joint positions or 6-DOF poses, which often induce excessive internal forces during contact, risking damage to both the robot and the environment~\cite{Zhao2023LearningFB, chi2023diffusion}. While end-effector force offers richer contact feedback, reliance on wrist force/torque (F/T) sensors limits scalability due to their high cost and scarcity~\cite{wang2013flexible}.

Recent efforts have attempted to incorporate force inputs into IL~\cite{kang2025robotic, beltran2020learning}, yet many treat force merely as an auxiliary signal for position control rather than leveraging it for compliant execution. Although emerging policies explore multimodal interaction~\cite{xue2025rdp, liu2025factr} and adaptive compliance~\cite{hou2025adaptive}, deploying such systems in low-cost, sensorless settings while ensuring stable, compliant torque-level execution remains an open challenge.

To address these issues, we present \textbf{FILIC}, a \textbf{F}orce-guided \textbf{I}mitation \textbf{L}earning framework with \textbf{I}mpedance torque \textbf{C}ontrol. First, we propose a dual-loop framework that integrates IL with impedance control. The Transformer-based IL policy serves as the outer loop, taking visual images, end-effector pose, and estimated force as observations to predict target poses, which are then fed into the inner-loop impedance controller to generate torque commands, forming a force-in, force-out dual-loop structure for compliant manipulation. Second, for scenarios without end-effector force/torque sensors, we develop a cost-effective end-effector force estimator by leveraging the mapping between the robot’s end-effector wrench and joint torques, based on the Jacobian matrix and a high-fidelity synchronized digital twin. Across simulation and real-world contact-rich tasks, FILIC with estimated end-effector force consistently outperforms both position-only and position-plus-joint-torque variants.

This work makes the following contributions:

\begin{figure*}[t]
    \vspace{5pt}
    \centering
    \includegraphics[width=\linewidth]{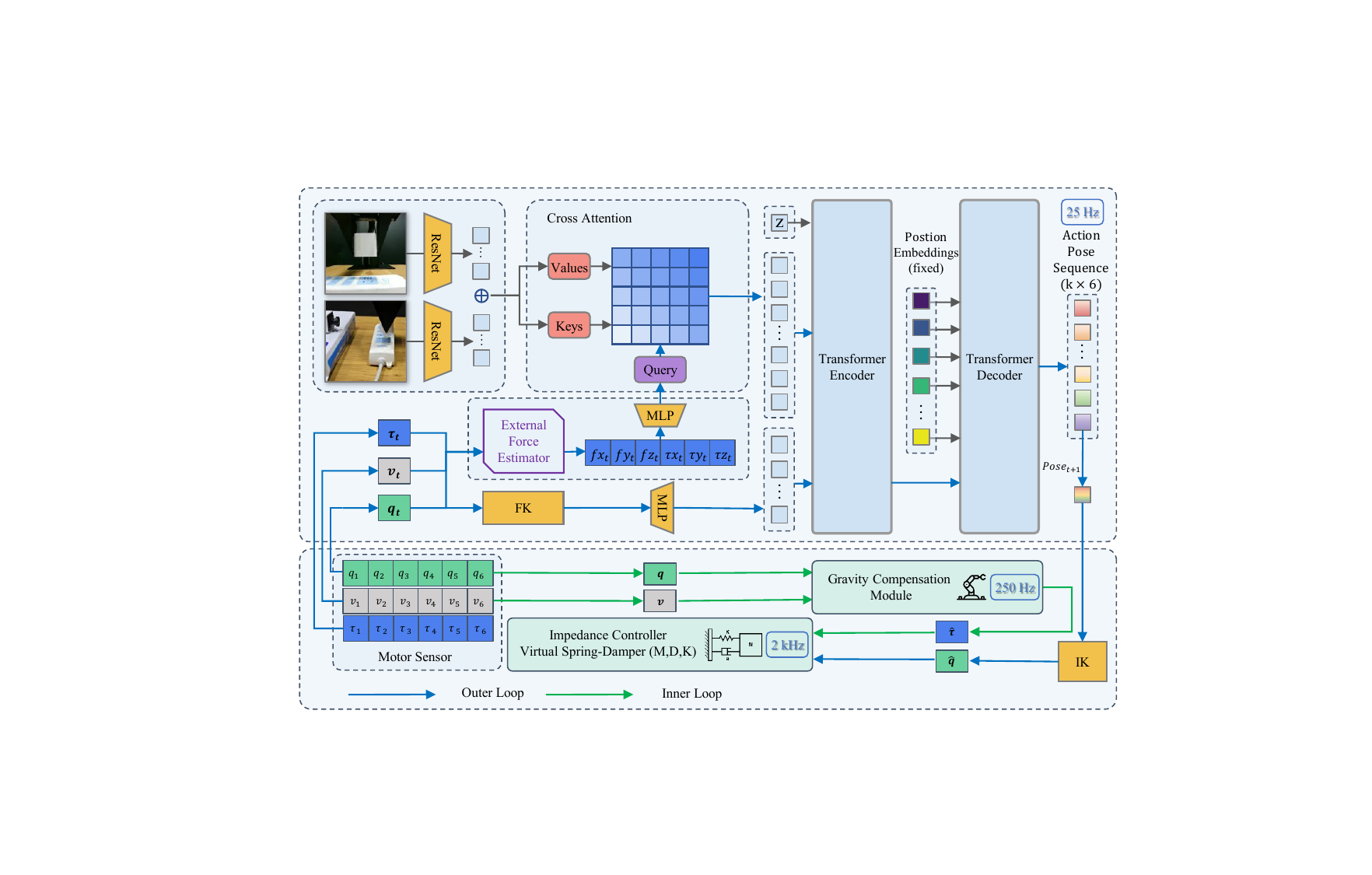}
    \caption{\textbf{Detailed architecture of FILIC}. \textbf{Outer loop:}  The external force estimator provides interaction forces, which are fused with forward kinematics (FK) outputs and further processed via MLP layers. Visual observations are encoded by dual ResNet backbones and integrated with estimated force embeddings through a cross-attention module. These multimodal features are passed into a standard Transformer architecture (e.g., ACT~\cite{Zhao2023LearningFB}) to generate pose sequences at 25 Hz. \textbf{Inner loop:} The predicted action pose is tracked through inverse kinematics (IK), while the inner control loop employs an impedance torque controller running at 2 kHz, supported by gravity compensation at 250 Hz.}
    \label{fig:main}
\end{figure*}

\begin{itemize}

\item A low-cost, sensorless method for estimating end-effector wrench using a synchronous digital twin, with an open-source implementation for accessibility.
\item A dual-loop framework combining an outer-loop Transformer-based imitation policy and an inner-loop impedance torque controller for compliant contact-rich manipulation with multimodal visual and force inputs.

\end{itemize}

%% file: Sections/2_RelatedWorks.tex
\section{Related Work}\label{sec 2}

\subsection{Imitation Learning}
Imitation learning (IL) seeks to acquire control policies directly from demonstrations, thereby circumventing the challenges of hand-crafted reward design but remaining prone to distribution shift under naive behavior cloning. Recent advances mitigate this issue through sequence- or chunk-level action prediction, which enhances temporal coherence and robustness~\cite{Zhao2023LearningFB}. In parallel, diffusion-based policy formulations recast action generation as a conditional denoising process, enabling the modeling of multimodal future behaviors~\cite{chi2023diffusion, zhao2024aloha}.
Beyond action representation, efforts to improve data efficiency and generalization have led to 3D point‑cloud–conditioned diffusion policies (DP3)~\cite{ze20243d} and vision–language–action (VLA) frameworks that integrate large-scale semantic priors~\cite{kim2024openvla}. Open-world variants such as $\pi0$ and $\pi0.5$ combine pretrained vision–language models (VLMs) with Flow Matching–based action experts to facilitate broader task transfer~\cite{black2410pi0,intelligence2025pi_}. In contrast, fully synthetic large-scale pretraining methods (e.g., GraspVLA) demonstrate competitive zero-shot sim-to‑real transfer~\cite{deng2025graspvla}.
Practical deployment has been further advanced through compact architectures (e.g., SmolVLA~\cite{shukor2025smolvla}), low-latency asynchronous chunk execution~\cite{black2025real}, and stabilization mechanisms such as Knowledge Insulating~\cite{driess2025knowledge}. Collectively, these developments highlight a convergence toward unified semantic reasoning and temporally consistent control~\cite{pan2024vision}. Nevertheless, current IL and VLA systems remain predominantly vision- and position-centric, leaving the exploitation of rich physical and interaction dynamics relatively underexplored.

\subsection{Force-Aware Robot Control and Learning}
Classical model-based strategies, such as impedance, admittance, and direct force control, establish principled mappings between motion and interaction forces, ensuring safe, compliant behavior~\cite{hogan1985impedance,abdullahi2024hybrid,vitrani2023improving} and stability across contact transitions in locomotion~\cite{cong2020contact}. However, in unstructured environments, modeling errors, hybrid contact dynamics, and the reality gap constrain purely analytical designs~\cite{kotha2024next}. While Deep Reinforcement Learning (DRL) combined with domain randomization has enabled agile and robust locomotion~\cite{margolis2024rapid} by leveraging dense proprioception and shaped rewards, manipulation presents distinct challenges: it requires proactive environmental modification~\cite{nguyen2019review}, suffers from severe visual sim-to-real transfer issues, demands multimodal sensing~\cite{hansen2022visuotactile}, and faces sparse reward structures~\cite{qi2025human}. Although augmenting perception with large Vision-Language Models (VLMs) or distilled demonstrations improves scene understanding~\cite{yu2024natural}, these approaches alone fail to yield force-sensitive compliance.

Recent efforts have sought to integrate explicit force awareness. Diffusion policies operating over 6D wrenches achieve precise contact reasoning but frequently omit visual inputs~\cite{wu2025tacdiffusion}. Similarly, variable impedance or admittance RL incorporates force-torque data for high-precision assembly yet often partially decouples vision~\cite{zhou2025admittance,zhou2025variable}. Hierarchical and residual formulations extend the action space with stiffness or damping parameters~\cite{tahmaz2025impedance,zhang2024residual,khazatsky2025droid}, while methods like SCAPE infer stiffness from position demonstrations to reduce expert burden; however, these remain position-centric and underutilize rich force signals~\cite{imicota,kim2022scape,luo2019reinforcement}. Systematic studies on torque fusion highlight the benefits of late fusion and auxiliary future-torque prediction for grounding internal representations~\cite{zhang2025ta}. Concurrently, frameworks such as Reactive Diffusion Policy and FACTR integrate multimodal contact cues with teleoperation interfaces~\cite{xue2025rdp,liu2025factr}, and the Adaptive Compliance Policy models task-dependent compliance within a diffusion-guided framework~\cite{hou2025adaptive}. However, these approaches predominantly assume sensor-rich setups, for example with wrist-mounted force/torque sensors, and their reliance on expensive hardware substantially limits both scalability and accessibility.

Collectively, these trends indicate a shift toward policies that couple semantic visual reasoning with learned, torque-aware control. Distinct from these sensor-rich or high-cost paradigms, FILIC targets resource-constrained settings by leveraging a sensorless, digital-twin-based wrench estimator to provide force-conditioned policy inputs, coupled with a fixed-impedance inner loop to ensure stable torque-level execution in contact-rich scenarios.

%% file: Sections/3_Preliminaries.tex
\section{Preliminaries}\label{sec 3}

\subsection{Imitation Learning}

Imitation Learning (IL) is a paradigm in which an agent learns to perform tasks by observing and mimicking expert demonstrations. Formally, given a dataset of state-action pairs $\mathcal{D} = \{ (s_i, a_i) \}_{i=1}^N$ collected from an expert policy $\pi^*$, the goal of IL is to learn a policy $\pi_\theta$ that minimizes the discrepancy between $\pi_\theta(a|s)$ and $\pi^*(a|s)$. Common approaches include Behavioral Cloning (BC), which treats IL as a supervised learning problem, and Inverse Reinforcement Learning (IRL), which infers a reward function underlying the expert’s behavior. IL is particularly valuable in settings where designing reward functions is difficult or when high-quality demonstrations are available.

\subsection{Dynamics and Wrench Propagation in Manipulators}

The dynamical model of a robotic manipulator is foundational for the analysis and control of interaction tasks. Given joint position $q$, velocity $\dot q$, and acceleration vectors $\ddot q$, the equations of motion are governed as follows: 
\begin{equation} \label{dynamic}
    \begin{aligned}
        M(q)\ddot{q} + C(q, \dot{q})\dot{q} + g(q) = \tau + \tau_{ext}
    \end{aligned}
\end{equation}

where $M(q)$, $C(q,\dot{q})$, and $g(q)$ are the inertia, Coriolis, and gravity terms, and $\tau$ is the applied joint torques. In the presence of external contact, the joint external torque $\tau_{ext}$ originates from a wrench $F_{ext}$ applied at the end-effector.

A fundamental mapping arises from the kinematics of constrained motion: the effective wrench $F_{ext}$ generated at the end-effector is algebraically related to the joint torques through the Jacobian transpose matrix $J(q)^T$:
\begin{equation} \label{forcemapping}
    \begin{aligned}
        \tau_{ext} = J(q)^T F_{ext}
    \end{aligned}
\end{equation}

This mapping is critical as it enables estimating external end-effector forces from joint current–derived torques without a dedicated force sensor. These inferred contact forces offer more intuitive physical information than raw joint torques, greatly improving a model's understanding of interaction force space.

\subsection{Impedance Control}
In robotic manipulation under unstructured environments, impedance control regulates the dynamic interaction between motion and external forces, which is typically modeled as a mass–spring–damper system, and can be implemented in either Cartesian or joint space. The desired joint torque $\tau^*$ is computed as:
\begin{equation} \label{idc}
    \begin{aligned}
        \tau^*=M(\ddot{q}_d-\ddot{q})+D(\dot{q}_d-\dot{q})+K(q_d-q)+\tau_{ff}
    \end{aligned}
\end{equation}
where $q_d,\dot{q}_d,\ddot{q}_d$ and $q,\dot{q},\ddot{q}$ are the desired and measured joint angles, velocities, and accelerations, respectively; $M,D,K$ are the joint-space inertia, damping, and stiffness matrices, and $\tau_{ff}$ is the feedforward compensation torque. 

During real-world deployment, acceleration measurements often suffer from significant errors, which can severely degrade control performance. As a result, the mass term is typically omitted, leading to a simplified impedance control formulation. Notably, such a impedance control can be realized through the MIT control mode of the actuators:
\begin{equation} \label{mit}
    \begin{aligned}
        \tau^*_{mit}=K_d(\dot{q}_d-\dot{q})+K_p(q_d-q)+\tau_{ff}
    \end{aligned}
\end{equation}
where $\tau^*_{mit}$ denotes the reference torque, while $K_p$ and $K_d$ denote the position and velocity gains, respectively.

%% file: Sections/4_Methodology.tex
\section{Method}\label{sec 4}

\subsection{Dual-Loop Imitation-Impedance Control} 

In this work, to achieve imitation-learning-based compliant manipulation, we design a dual-loop control architecture that integrates a high-level imitation learning policy with a low-level impedance controller, as illustrated in Fig.~\ref{fig:main}. The outer loop employs imitation learning to generate adaptive task-level motion commands from multimodal observations, while the inner loop enforces stable and compliant execution through impedance control at the torque level.

\textbf{Outer-loop imitation learning policy.}
We employ a Transformer-based model as the outer-loop imitation learning policy as shown by the blue flow in Fig.~\ref{fig:teaser} and Fig.~\ref{fig:main}.

For state estimation, joint positions $[q_1,...,q_n]_t$, velocities $[v_1,...,v_n]_t$, and torques $[\tau_1,...,\tau_n]_t$ are obtained from the robot’s internal sensors and converted via forward kinematics and dynamics into the end-effector Cartesian pose and the corresponding external wrench $F_{ext}=[fx_t,fy_t,fz_t,\tau x_t, \tau y_t, \tau z_t]$. This proprioceptive information is synchronized with RGB images from two external cameras, all down-sampled to 25 Hz for efficient processing. Each visual stream is encoded using a ResNet backbone, while the force vector is mapped through a multilayer perceptron (MLP). The resulting visual and force features are then fused via a cross-attention module. This fused representation is subsequently concatenated with two additional components: (1) the end-effector pose features, also projected through an MLP, and (2) a CVAE~\cite{cvae} latent style variable $Z$, which is derived from encoding the action pose sequence and current end-effector pose. Notably, $Z$ is omitted during inference~\cite{Zhao2023LearningFB}.

The Transformer-based model network outputs a sequence of target 6-DOF end-effector poses at 25 Hz, following the action chunking with temporal ensemble processing to enhance robustness and smoothness inspired by Action Chunking with Transformers (ACT)~\cite{Zhao2023LearningFB}. Except for the perception input module, the detailed architecture of the other modules follows ACT. These high-level references are passed to the inner-loop impedance controller, which translates them into compliant torque-level commands, thereby ensuring both task fidelity and safe interaction.

\textbf{Inner-loop impedance controller.}
As depicted by the green flow in Fig.~\ref{fig:teaser} and Fig.~\ref{fig:main}, to ensure robust and stable control on the physical platform, we assume the tasks are approximately quasi-static and thus simplify the impedance model by excluding the inertial component, while still modeling damping effects. The resulting formulation retains only the spring–damper terms, which balance compliance with stability.

At each control cycle, the imitation learning policy provides a 6-DOF Cartesian target pose $pose_{t+1}$ at 25 Hz. This target is mapped into desired joint angles $\hat{q}$ through real-time inverse kinematics. In parallel, joint positions $q$ and velocities $v$ are acquired from the robot’s internal sensors at a higher rate of 250 Hz. These high-frequency proprioceptive signals are integrated into a joint-space impedance control law that computes torque commands $\tau_{cmd}$ at 2 kHz formulated as:
\begin{equation} \label{idc_ours}
    \begin{aligned}
        \tau_{cmd}=B(\hat{v}-v)+K(\hat{q}-q)+\hat {\tau}
    \end{aligned}
\end{equation}
where $K$ and $B$ are the preset stiffness and damping coefficients, $\hat{v}$ is typically set to 0, and $\hat {\tau}$ is the feedforward compensation torque (considering both gravity and Coriolis forces), computed at 250 Hz. According to Equation~(\ref{mit}) and ~(\ref{idc_ours}), $\hat{v}, \hat{q}, \hat {\tau}$ can be directly provided as inputs to the motor control function in MIT mode, from which the desired torque to be executed is computed. We use fixed $(K,B)$ to prioritize stable deployment, leaving policy-conditioned variable impedance for future work. The three loop rates are chosen to balance perception cost against control stability: the policy runs at 25 Hz, matching the camera frame rate and the down-sampled demonstration stream while keeping Transformer inference with temporal ensembling tractable; the gravity and Coriolis compensation $\hat{\tau}$ is refreshed at 250 Hz, which is sufficient to track configuration-dependent dynamics that vary only with relatively slow arm motion; and the impedance law executes at 2 kHz to provide the high feedback bandwidth required for stable, compliant torque tracking during contact.

In summary, the proposed hierarchical structure uniquely leverages end-effector forces as inputs for imitation learning and controls joint torques as outputs through impedance control, achieving robust, compliant, and demonstration-informed force control in contact-rich tasks.

\subsection{External Force Estimation via Digital Twin}

\textbf{Jacobian-based external force estimation.}
To exploit the intuitive and decoupled information provided by end-effector forces while overcoming the lack of such sensors in most affordable commercial robotic arms, we reconstruct the 6-dimensional end-effector wrench by mapping joint torques through the inverse relationship defined by the Jacobian, as presented the dark purple modules in Fig.~\ref{fig:teaser} and Fig.~\ref{fig:main}. The estimated force is then employed as perceptual input for imitation learning. Based on the mapping relationship between forces and torques presented in Equation~(\ref{forcemapping}), we obtain the following estimation formula for the external end-effector force:
\begin{equation} \label{fext}
    \begin{aligned}
        F_{ext}=J(q)^{T+}(\tau_{obs}-\tau_{inv})
    \end{aligned}
\end{equation}
where $\tau_{obs}$ and $\tau_{inv}$ are the observed and expected dynamics-based joint torques, respectively. The pseudo-inverse $J(q)^{T+}$ is calculated via SVD to ensure stable least-squares approximation even with dimensional misalignment.

\textbf{Synchronous digital twin.} Moreover, to robustly compute expected joint torques and estimate end-effector forces in real-world robots affected by factors such as sensor noise, joint friction, and mechanical compliance, we employ a synchronous digital twin framework. Specifically, a high-fidelity robotic arm model is instantiated in the MuJoCo simulator, where measured joint positions, velocities, and torques from the physical robot are continuously synchronized in real time. Exploiting MuJoCo’s precise dynamics simulation, the framework computes expected joint torques $\tau_{inv}$ and infers the corresponding external wrench $F_{ext}$ on the end-effector. This approach obviates the need for additional high-precision force or acceleration sensors, minimizes hardware dependency, and can be generalized to different robotic platforms by simply updating the simulation model.

\subsection{Demonstration Data Collection with Vibro-Haptic Feedback}

Collecting high-quality demonstration data is essential for imitation learning, particularly when force signals are included as part of the model input. To ensure that demonstrations capture informative variations in force, we use a low-cost handheld teleoperation interface that enhances the operator’s perception of interaction forces and improves the quality of recorded trajectories.

Our framework employs a handheld controller with vibrotactile haptic feedback. Estimated end-effector forces, computed from Equation (6), are mapped to haptic vibration cues, where vibration frequency increases discretely as force magnitude increases. This design conveys qualitative information about contact intensity, enabling demonstrators to intuitively adjust their actions. As a result, the collected trajectories tend to better capture meaningful force variations, thereby improving the robustness of the learned IL policy.

%% file: Sections/5_Experiments.tex
\section{Experiments}
We evaluate FILIC from three complementary perspectives. We first quantify the accuracy of the external-force estimator, then examine qualitative force/torque traces in representative contact-rich tasks, and finally report policy-level ablations in both simulation and real-world settings.

\begin{figure*}[t]
    \vspace{5pt}
    \centering
    \includegraphics[width=\linewidth]{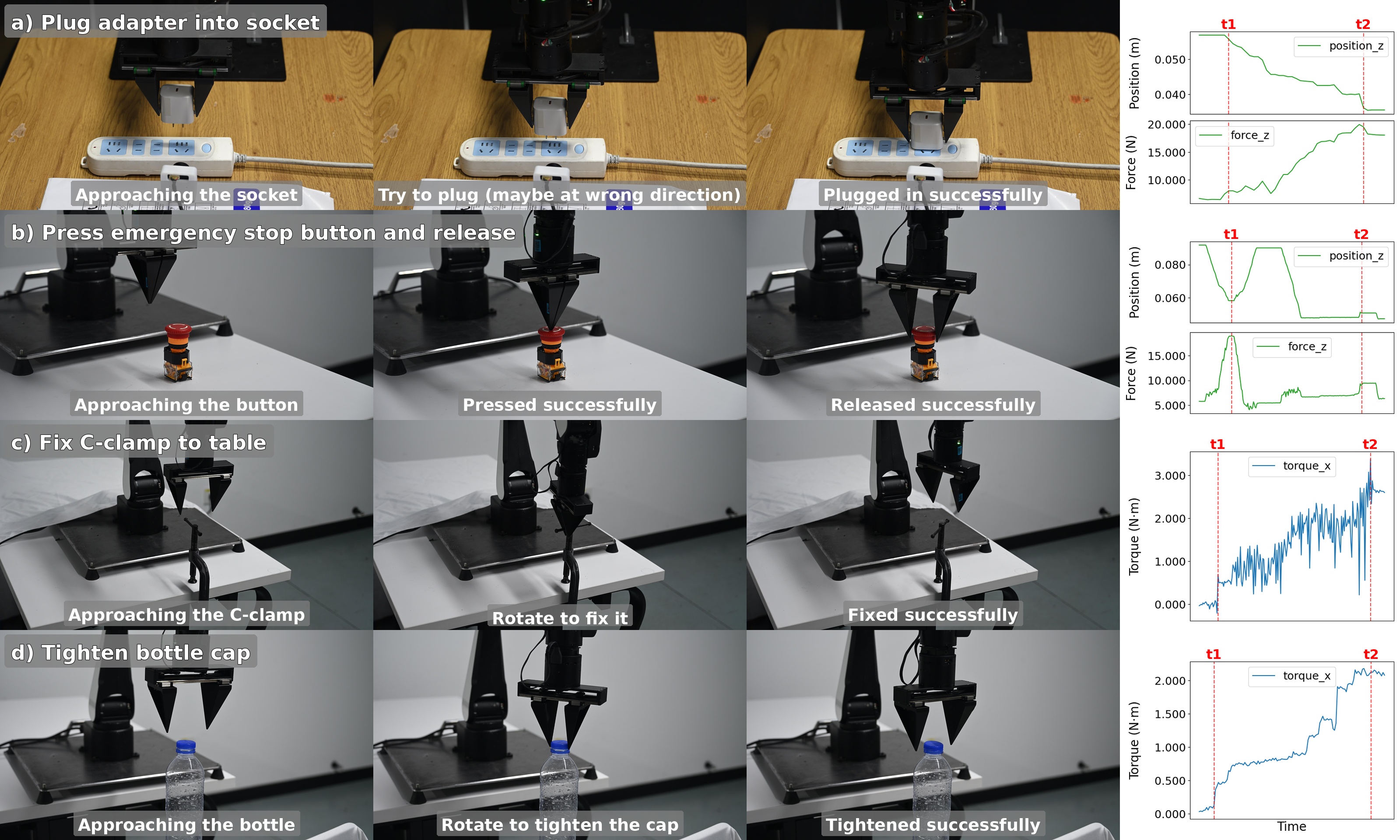}
    \caption{\textbf{Process and force feedback of four demonstration tasks.} Peaks or sudden shifts delineate key contact events, such as pressing, releasing, locking, or tightening, capturing the diverse physical interactions inherent in the manipulation sequence.}
    \label{fig:placeholder}
\end{figure*}

\subsection{Quantitative Accuracy of External Force Estimation}

\begin{figure}[t]
    \vspace{5pt}
    \centering
    \includegraphics[width=\linewidth]{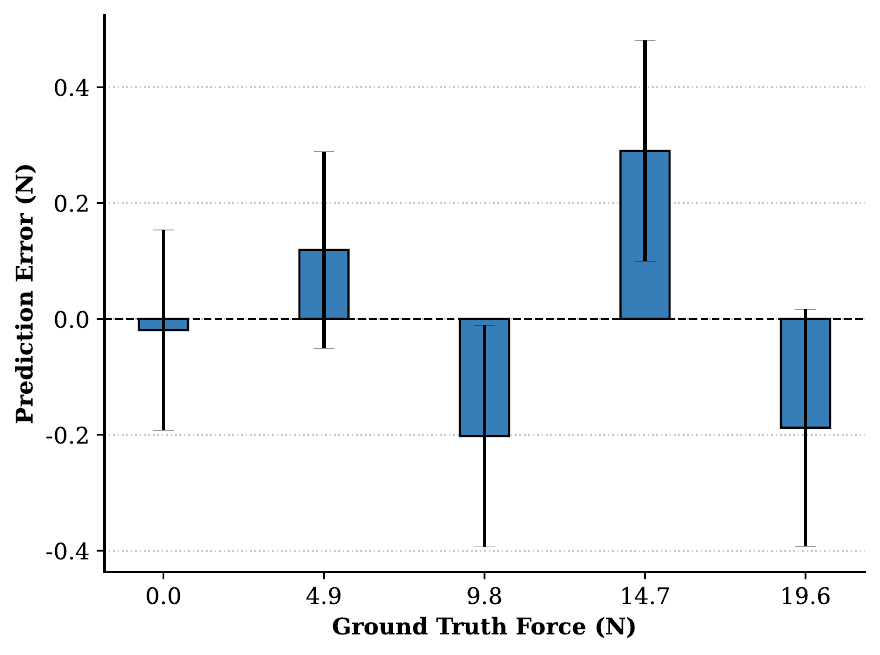}
    \caption{\textbf{Prediction error of the external-force estimator under different static loads.} Different known masses are attached to the end-effector, corresponding to true downward forces under gravitational acceleration $g=9.81$ m/s$^2$. }
    \label{fig:pred_error}
\end{figure}

We evaluate the absolute accuracy of the external-force estimator with a controlled static-load test on a 6-DOF AIRBOT Play manipulator. For each load condition, the robotic arm is moved to multiple poses and held stationary, and 30 repeated estimates are recorded. The prediction error is defined as the difference between the estimated Z-axis external force and the true gravitational load. Fig.~\ref{fig:pred_error} reports the mean and standard deviation of this error for each load.

\subsection{Qualitative Analysis of Estimated Force Traces}
We present real-world force and torque traces for four contact-rich tasks to illustrate how interaction events appear in the estimated signals. Fig.~\ref{fig:placeholder} shows representative trajectories.

\textbf{Plug adapter into socket.}
The socket is fixed on the table, and the robotic arm inserts the adapter along the insertion direction. The Z-axis force rises at $t_1$ and continues increasing until a peak at $t_2$.

\textbf{Press emergency stop button and release.}
For the emergency stop button, the robotic arm first presses downward, then twists to release. A pronounced Z-axis force peak appears near $t_1$ during pressing, followed by a sharp change at $t_2$ during spring-driven release.

\textbf{Fix C-clamp to table.}
In the C-clamp task, wrist rotation produces a rapid rise in X-axis torque at $t_1$. The torque then increases gradually and reaches a maximum around $t_2$ as tightening proceeds.

\textbf{Tighten bottle cap.}
The bottle-cap trace shows a similar pattern: X-axis torque increases abruptly at $t_1$, then climbs progressively and peaks at $t_2$ during final tightening.

Across these tasks, key interaction stages align with local extrema or abrupt transitions in force/torque trajectories, providing directly observable contact cues.

\subsection{Policy Evaluation Setup}

We evaluate FILIC in simulation and on a real robotic arm. The setup is designed as a controlled ablation in which all model components and hyperparameters are fixed, and only the proprioceptive contact cue is changed.

\subsubsection{Simulation Setup}
\begin{figure}[t]
    \vspace{5pt}
    \centering
    \includegraphics[width=\linewidth]{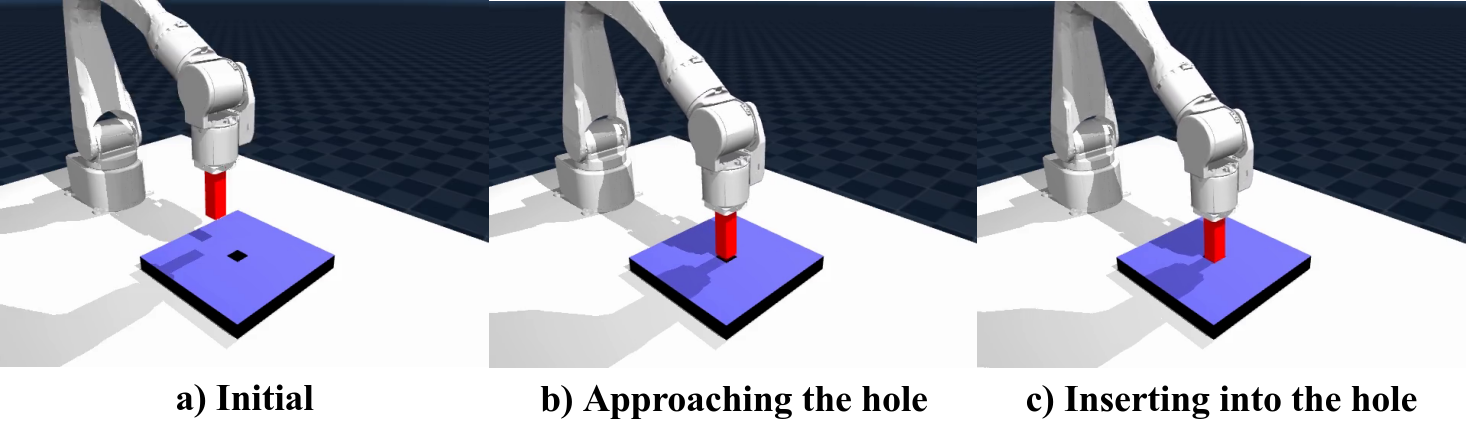}
    \caption{\textbf{Simulation scene setup and step-by-step process of the peg-in-hole task.} The panels show (a) the initial configuration, (b) approaching the hole, and (c) inserting the peg into the hole.}
    \label{fig:sim_seup}
\end{figure}

We use MuJoCo for contact-rich manipulation. The simulated robot is a 6-DOF manipulator with a fixed rectangular peg (0.95 cm side length). The environment contains a flat tabletop and a 1 cm square hole. Two RGB cameras provide local and global observations. The simulation scene and the step-by-step task progression are shown in Fig.~\ref{fig:sim_seup}.

\textbf{Task.} The simulation task is peg-in-hole insertion. The end-effector must align the rectangular peg with the square hole and complete insertion. End-effector orientation is fixed perpendicular to the table, so only translational DoFs $(x, y, z)$ are controlled. At episode start, the hole is slightly below and in front of the peg.

\begin{figure}[t]
    \vspace{5pt}
    \centering
    \includegraphics[width=\linewidth]{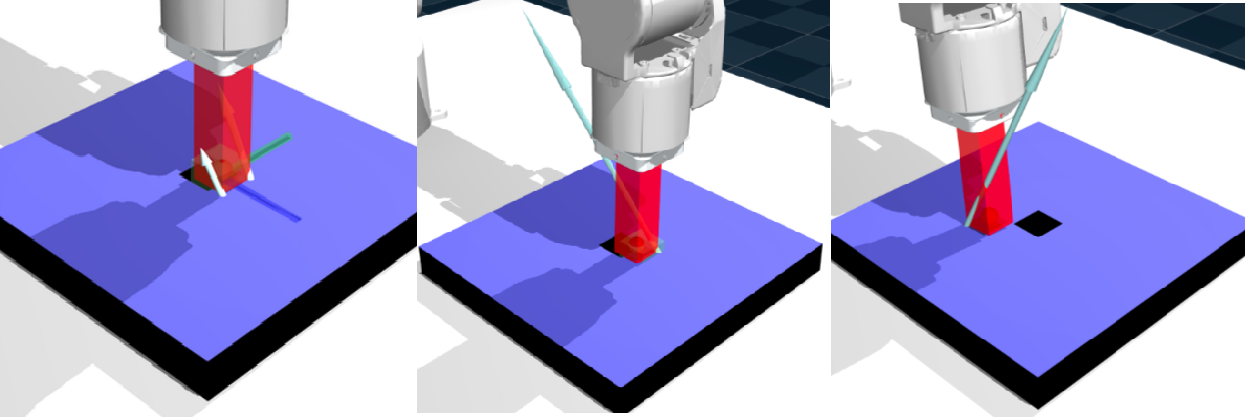}
    \caption{\textbf{Visualization of contact forces in the simulator.} The direction and length of the blue arrow represent the direction and relative magnitude of the force at the contact point.}
    \label{fig:sim_force}
\end{figure}

\textbf{Data collection.} Demonstrations are collected with the haptic-feedback handheld controller to include corrective behavior under contact. In addition, contact forces are visualized in the simulator to provide more intuitive feedback, as shown in Fig.~\ref{fig:sim_force}. Before each trajectory, the end-effector is reset to a fixed initial pose and the hole center is perturbed by a random 1 mm offset. The dataset contains 150 trajectories: 50 contact-free single-shot insertions and 100 trajectories with force-guided corrections. Each trajectory includes synchronized RGB images, end-effector Cartesian position, joint torques, estimated end-effector forces, and operator Cartesian commands.

\subsubsection{Real-World Setup} The robotic arm is fixed on a desk. Two RGB cameras are placed in front of and to the left of the target object for scene observation.

\textbf{Tasks.} We use the four tasks shown in Fig.~\ref{fig:placeholder}. At initialization, the end-effector is above the object with an approximate lateral offset of 2 cm.

\textbf{Data collection.} Demonstration data are collected with the same haptic-feedback handheld controller. For each task, the dataset contains 30 trajectories. For the adapter insertion task, we collect 10 contact-free single-shot insertions and 20 trajectories with force-guided corrective motions. For the emergency stop button operation task, the execution logic differs slightly from Fig.~\ref{fig:placeholder} b). The button starts in a random ON or OFF state. The arm first rotates the button toward the ON state. If the initial state is OFF, the arm then presses to return it to OFF; if the initial state is ON, it remains ON after release. A trial is counted as failed when the final state differs from the initial state. For C-clamp fixation task and bottle cap tightening task, the initial tightness is randomized, and the final condition requires a clearly tightened, non-loose state; otherwise the trial is marked as failed. Only successful demonstrations are retained. Data modalities are the same as in simulation.

\begin{table}[h]
\vspace{5pt}
\centering
\caption{\textbf{Model hyperparameters in the experiment.}}
\label{tab:hyper}
\normalsize
\begin{tabular}{l|c}
\hline
Hyperparameter & Value \\
\hline
Learning rate & 2e-5 \\
Batch size & 16 \\
\# Encoder layers & 4 \\
\# Decoder layers & 7 \\
Feedforward dimension & 3200 \\
Hidden dimension & 512 \\
\# Heads & 8 \\
Chunk size & 25 \\
Beta & 10 \\
Dropout & 0.1 \\
\hline
\end{tabular}
\end{table}

\subsection{Policy Evaluation Results}

We compare three proprioceptive settings in both simulation and real-world tasks: end-effector (EE) position only, EE position + joint torque, and EE position + estimated EE force. Model architecture and training hyperparameters are identical across variants (Table~\ref{tab:hyper}). All settings use RGB observations and EE position; only the additional contact cue differs. Each simulation condition is evaluated over 50 trials, while each real-world condition uses 30 trials. Results are reported in Tables~\ref{tab:success_sim} and \ref{tab:success_real}.

Three observations follow from these results. First, the position-only baseline reaches 68.0\% in simulation and 26.6\%--46.7\% across real-world tasks, and the dominant failure cases are consistent across tasks: in peg/adapter insertion task, once the peg/adapter is laterally misaligned, the policy often cannot autonomously recover; in E-stop task, the final press-to-OFF action is inconsistent and appears largely random regardless of the initial ON/OFF state; in Clamp task and Cap task, the policy frequently fails to complete tightening when the initial state is relatively loose. These failures indicate that position-only observations miss key task-state information needed for reliable contact-rich control.

Second, adding force-related proprioception improves success rates in all settings. In simulation, the gain over position-only input is +12.0 pp with joint torque (80.0\%) and +22.0 pp with estimated EE force (90.0\%). On real-world tasks, estimated EE force improves over position-only input by +33.3 pp (Plug), +40.0 pp (E-stop), +46.6 pp (Clamp), and +53.4 pp (Cap).

\begin{table}[t]
\vspace{5pt}
\centering
\caption{\textbf{Simulation success rate under different proprioceptive inputs.} All models take end-effector position and RGB images as inputs; they differ only in the additional proprioceptive cue.}
\label{tab:success_sim}
\normalsize
\begin{tabular}{l|c}
\hline
Proprioceptive Input & Success Rate (\%) \\
\hline
EE position & 68.0 \\
EE position + joint torque & 80.0 \\
EE position + EE force & 90.0 \\
\hline
\end{tabular}
\end{table}

\begin{table}[t]
\vspace{5pt}
\centering
\caption{\textbf{Real-robot success rate (\%) across four tasks.} Input abbreviations: Pos = EE position, JT = joint torque, EF = estimated EE force. Task abbreviations: Plug = plug adapter into socket, E-stop = operate emergency stop button, Clamp = fix C-clamp to table, Cap = tighten bottle cap.}
\label{tab:success_real}
\normalsize
\begin{tabular}{l|c|c|c}
\hline
Task & Pos & Pos + JT & Pos + EF \\
\hline
Plug & 46.7 & 63.3 & 80.0 \\
E-stop & 43.3 & 60.0 & 83.3 \\
Clamp & 30.0 & 63.3 & 76.6 \\
Cap & 26.6 & 70.0 & 80.0 \\
\hline
\end{tabular}
\end{table}

Third, estimated EE force consistently outperforms raw joint torque. In simulation, success increases from 80.0\% to 90.0\% (+10.0 pp). In real-world tasks, the improvement is +16.7 pp (Plug), +23.3 pp (E-stop), +13.3 pp (Clamp), and +10.0 pp (Cap). These differences indicate that the estimated EE force provides a more task-aligned contact signal than joint torque under the same policy architecture and training setup.

%% file: Sections/6_Conclusion.tex
\section{Conclusion}

We present FILIC, a dual-loop force-aware imitation learning framework that unifies a high-level Transformer policy with a low-level impedance controller for compliant, contact-rich manipulation. FILIC leverages a sensorless, Jacobian-based end-effector force estimation, fused with RGB observations, while the inner impedance loop ensures torque-level compliance. Across simulation and real-world contact-rich tasks, replacing raw joint torque with estimated end-effector force consistently improves policy performance. These results show that explicit force representation and inner-loop compliance complement modern IL policies without requiring expensive wrist F/T sensors.

\textbf{Future work}. Our sensorless force reconstruction relies on modeled dynamics and uses fixed impedance parameters. Future work will integrate variable impedance for dynamic compliance in complex interactions, enhance force estimation robustness against modeling inaccuracies, and explore implicit force reasoning.

%% file: Sections/7_Bibliography.tex
\balance

\bibliographystyle{ieeetr}
\bibliography{root}